\documentclass{article}
\usepackage{graphicx} 
\usepackage{siunitx}
\usepackage{booktabs}
\usepackage{amsmath} 
\usepackage{amssymb}  
\usepackage{hyperref}
\hypersetup{
  colorlinks   = true, 
  urlcolor     = blue, 
  linkcolor    = blue, 
  citecolor   = red 
}

\usepackage{caption}
\usepackage{subcaption}

\usepackage[style=numeric]{biblatex}
\usepackage{listings}
\usepackage{xcolor}
\usepackage{soul}

\definecolor{codegreen}{rgb}{0,0.6,0}
\definecolor{codegray}{rgb}{0.5,0.5,0.5}
\definecolor{codepurple}{rgb}{0.58,0,0.82}
\definecolor{backcolour}{rgb}{0.95,0.95,0.92}

\lstdefinestyle{mystyle}{
    backgroundcolor=\color{backcolour},   
    commentstyle=\color{codegreen},
    keywordstyle=\color{magenta},
    numberstyle=\tiny\color{codegray},
    stringstyle=\color{codepurple},
    basicstyle=\ttfamily\footnotesize,
    breakatwhitespace=false,         
    breaklines=true,                 
    captionpos=b,                    
    keepspaces=true,                 
    numbers=left,                    
    numbersep=5pt,                  
    showspaces=false,                
    showstringspaces=false,
    showtabs=false,                  
    tabsize=2
}
\usepackage[margin=0.5in]{geometry}
\usepackage{array}
\newcolumntype{H}{>{\setbox0=\hbox\bgroup}c<{\egroup}@{}}
\usepackage[inkscapeformat=png]{svg}

\DeclareMathOperator*{\argmin}{arg\,min}

\title{Deriving The Fundamental Equation of Earthmoving and Configuring Vortex Studio Earthmoving Simulation for Soil Property Estimation Experimentation}
\author{W. Jacob Wagner$^{1,2,*}$}

\date{%
        $1$Department of Electrical and Computer Engineering, University of Illinois at Urbana-Champaign, Urbana, IL, United States\\%
        $2$Construction Engineering Research Laboratory, Engineer Research and Development Center, United States Army Corps of Engineers, Champaign, IL, United States\\%
        $*$Corresponding author: wjwagne2@illinois.edu
    }

\begin{document}

\maketitle

This document serves as supplementary material to the ISTVS papers \cite{wagner2023,wagner2025}.
It covers the derivation of the fundamental equation of earthmoving for a flat blade moving through sloped soil and provides some information regarding the advanced configuration of Vortex Studio's soil-tool interaction simulation.

\section{Soil Mechanics and the Fundamental Equation of Earthmoving (FEE)}
Soil mechanics is complex and no single theory is suitable for analyzing all scenarios that may be of interest to engineers.
Critical state theory is useful for characterizing the strength of saturated soils while the Mohr-Coulomb theory is more suitable for drained soils.
Additionally, penetration of soil is modeled differently than cutting of soil.
For cutting soil, a commonly made assumption is that failure surfaces are planar, although it is more accurate to describe this using a logarithmic spiral \cite{Mckyes1965_ch3}.
Discrete element models provide the most accurate representation of the true soil physics, but are computationally expensive.
In the work described in this thesis, we assume the Mohr-Coulomb model of soil strength.

\subsection{The Mohr-Coulomb Soil Model}
Coulomb developed a model of soil shear strength in 1776 describes soil not just having a static/cohesive shear strength as solid materials do, but additionally having a component that varies with the normal force i.e. a frictional component.
\begin{equation} \label{eq:coulomb-soil}
    \tau = c + \sigma_n \text{tan}(\phi)
\end{equation}
where $\tau$ is the soil shear strength (maximum shear stress), c is the cohesion, $\sigma_n$ is the normal stress on the sliding plane, and $\phi$ is the angle of internal friction \cite{Mckyes1989_ch2}.
So, unlike a purely cohesive material, soil with a significant angle of internal friction will shear at a different point depending on the applied normal stress.
Cohesive soils have very little friction such as undrained saturated fine grained soil, whereas frictional soils have very little cohesion e.g. dry sand.
In addition to soil type, $c$ and $\phi$ vary significantly with soil moisture content and the soil compaction.

\begin{figure}[t]
    \centering
    \includegraphics[width=0.75\textwidth]{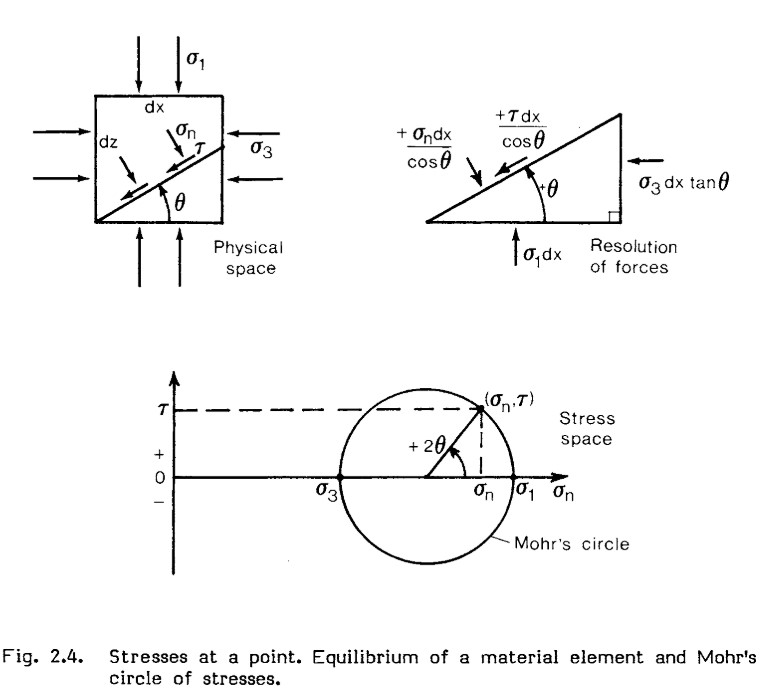}
    \caption{Mohr's circle diagram reproduced from \cite{Mckyes1989_ch2}.}
    \label{fig:mohrs-circle}
\end{figure}

In order to compute the soil shear strength $\tau$ the normal stress acting on the failure plane must be known.
In many cases, knowledge of the forces/stresses present is available, but they are often acting on different planes than the failure plane.
By analyzing the stresses acting on a rectangular finite element of the soil and balancing the forces along the new desired plane at an angle $\theta$ with respect to the original the equations for Mohr's circle can be derived, see Figure \ref{fig:mohrs-circle}.
This approach assumes that the system is in static equilibrium, i.e. no acceleration \cite{Mckyes1989_ch2}.

\begin{figure}[h]
    \centering
    \includegraphics[width=0.75\textwidth]{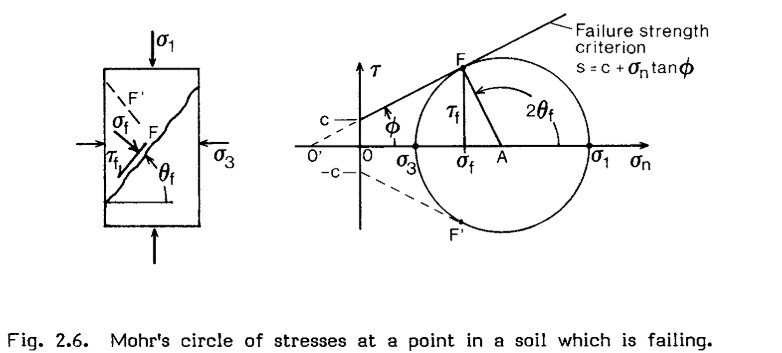}
    \caption{The failure strength criterion line is tangent to Mohr's circle due to our assumption of no acceleration. If the line did not touch the circle, then the failure condition would not be met. If the line touched the circle at more than one point, then the the stress at some plane within the material would be greater than the failure stress meaning that the soil would have to be accelerating due to a force imbalance. The angle $\sigma_f$ is the angle of the soil failure plane from the principal stress plane. Reproduced from \cite{Mckyes1989_ch2}.}
    \label{fig:mohr-coulomb}
\end{figure}

Applying Mohr's circle to the stresses for the Coulomb soil failure condition, Equation \ref{eq:coulomb-soil}, the relationship between the failure stresses and stresses applied to the material at different angles can be defined, see Figure \ref{fig:mohr-coulomb}.
This leads to the Mohr-Coulomb model of soil failure.
This analysis can be extended to determine the angle of the failure plane with respect to the horizontal as well.

\subsection{Derivation of the FEE for a Bulldozer Blade in Sloped Terrain}
\begin{figure}[h]
    \centering
    \includegraphics[width=0.6\linewidth]{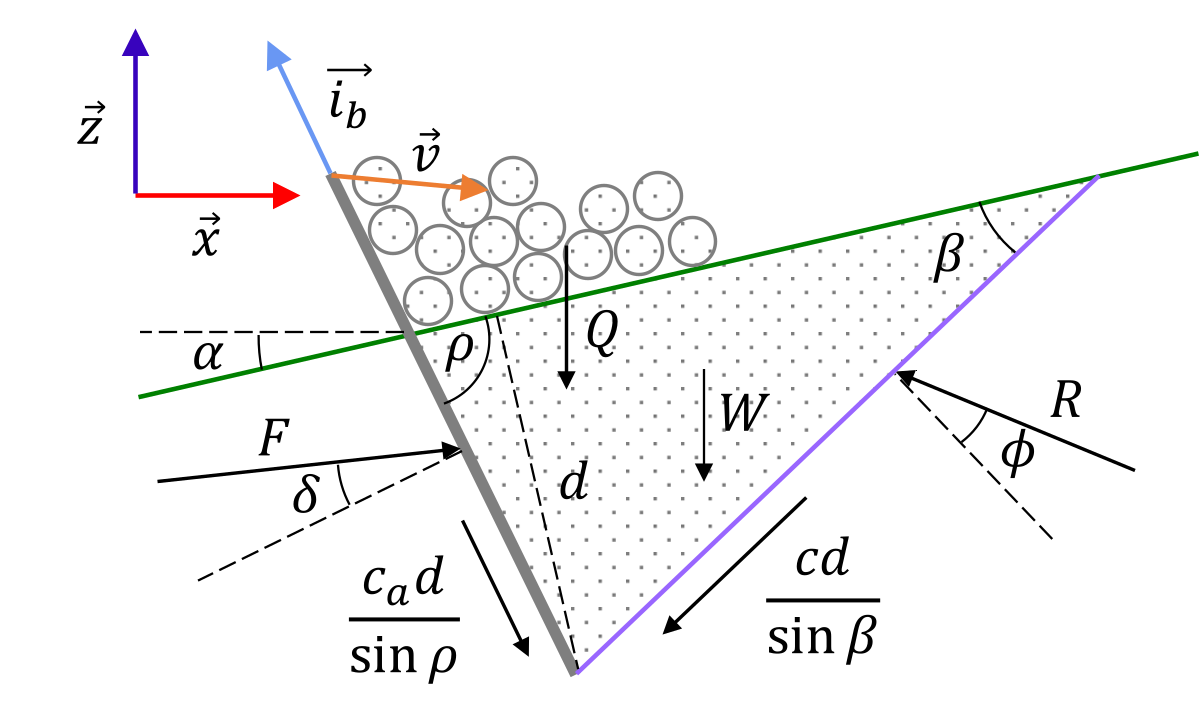}
    \caption{FEE forces nomenclature diagram}
    \label{fig:FEE_nomenclature}
\end{figure}

The fundamental equation of earthmoving (FEE) model is a powerful tool for predicting the cutting force required to move soil with a blade. In this section, the FEE model for sloped terrains, as presented by \cite{Holz2013} will be derived.
First assume a Mohr-Coulomb model of homogeneous soil where the blade moves through the soil primarily horizontally in the direction.
The soil is assumed to shear along a plane indicated by the blue line in Figure \ref{fig:FEE_nomenclature}.  
A flat blade is moving in the direction $V$ causes the soil to fail along this surface generating a wedge of soil as shown in Figure \ref{fig:FEE_nomenclature}.
Assuming the passive failure case \cite{Mckyes1989_ch6}, i.e. as the blade moves forward the wedge moves up along the blade and to the right along the soil failure line.
Without the use of Mohr's circle, Coulomb, and later Mckyes, demonstrated that by invoking the condition of static equillibrium, it is possible to derive an expression for the cutting force per blade width $F$ \cite{Mckyes1989_ch6, Mckyes1965_ch3}.
What follows is a step-by-step derivation of the FEE for a flat bulldozer blade moving through a sloped terrain.
The textbook ``Agricultural Engineering Soil Mechanics" by E. Mckyes is heavily used as a reference throughout this section and provides a good introduction to the subject of soil mechanics for the purposes of this work.

\begin{figure}[h]
    \centering
    \includegraphics[width=0.6\linewidth]{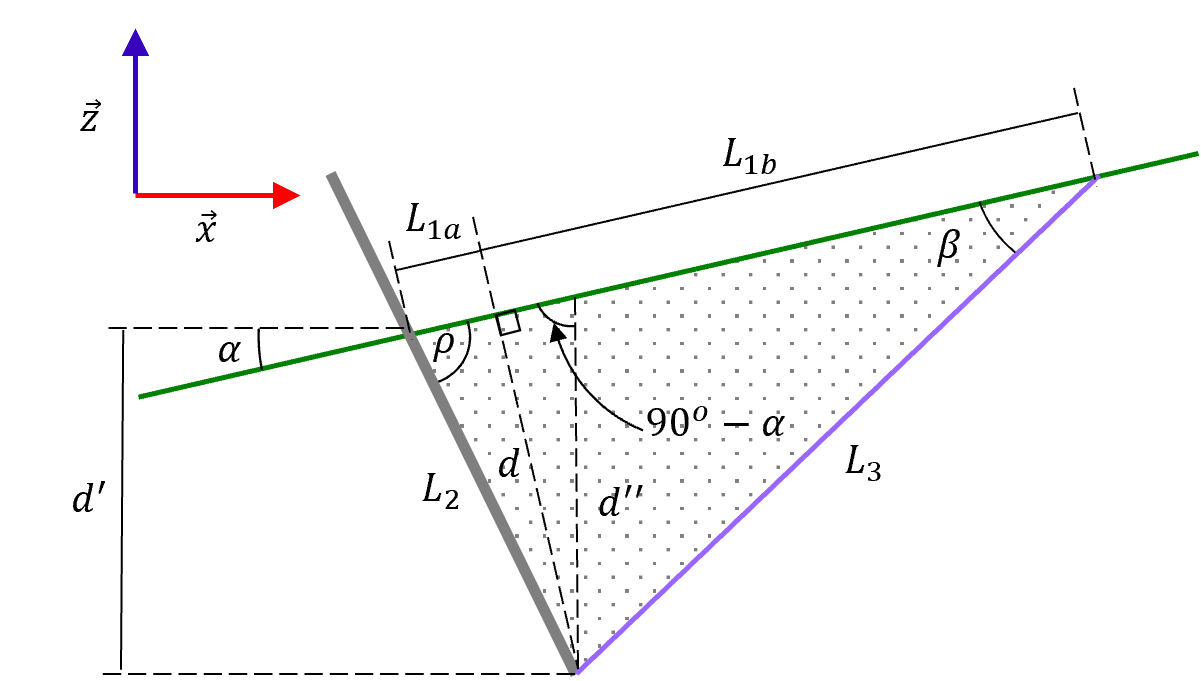}
    \caption{FEE geometry nomenclature diagram}
    \label{fig:FEE_measurements}
\end{figure}

First let us define some of the geometric parameters in terms of the main parameters $d$, $\rho$, and $\beta$
\begin{align}
    \begin{matrix}
        L_1 = L_{1a} + L_{1b}, & d^{\prime} = \frac{d \sin(\rho - \alpha)}{\sin(\rho)}\\
        L_{1a} = d \cot(\rho), &  L_2 = \frac{d}{\sin(\rho)}\\
        L_{1b} = d \cot(\beta), & L_3 = \frac{d}{\sin(\beta)}\\
        d^{\prime\prime} = \frac{d}{\cos(\alpha)}
    \end{matrix}
\end{align}

\begin{table}
    \caption{Nomenclature. Note the following abbreviations are used: unit blade width (ubw.), with respect to (wrt.)}
    \centering
    \begin{tabular}{@{}llH@{}}
        \toprule
        Parameter (Units) & Description & Derived Status\\
        \midrule
        \multicolumn{3}{c}{Geometric Parameters}\\
        \midrule
        $\rho$ (\unit{\radian}) & Blade angle wrt. surface & Required \\
        $\alpha$ (\unit{\radian}) & Surface angle & Required \\
        $\beta$ (\unit{\radian}) & Soil failure angle wrt. surface & Derived via optimization \\
        $d$ (\unit{\meter}) & Cut depth perpendicular to surface & Required \\
        $w$ (\unit{\meter}) & Blade width & Required \\
        $d^{\prime}$ (\unit{\meter}) & Cut depth at ground int. wrt. horizontal & Derived \\
        $d^{\prime\prime}$ (\unit{\meter}) & Cut depth at blade tip wrt. horizontal & Derived \\
        $L_1$ (\unit{\meter}) & Soil wedge base length & Derived \\
        $L_{1a}$ (\unit{\meter}) & Component of $L_1$ & Derived \\
        $L_{1b}$ (\unit{\meter}) & Component of $L_1$ & Derived \\
        $L_2$ (\unit{\meter}) & Blade-soil contact length & Derived \\
        $L_3$ (\unit{\meter}) & Soil failure surface length & Derived \\
        $\Delta V$ (\unit{\meter\squared}) & Swept volume per ubw. &  Required? \\
        \midrule
        \multicolumn{3}{c}{Soil Properties}\\
        \midrule
        $\phi$ (\unit{\radian}) & Soil internal angle of friction & Required \\
        $\delta$ (\unit{\radian}) & Soil-tool friction angle & Required \\
        $c_a$ (\unit{\pascal}) & Soil-tool adhesion & Required \\
        $c$ (\unit{\pascal}) & Soil cohesion & Required \\
        $\gamma$ (\unit{\newton\per\meter\cubed}) & Soil moist unit weight & Required \\
        \midrule
        \multicolumn{3}{c}{Forces}\\
        \midrule
        $Q$ (\unit{\newton\per\meter}) & Surcharge force per ubw. & Derived\\
        $q$ (\unit{\pascal}) & Surcharge pressure & Required?\\
        $R$ (\unit{\newton\per\meter}) & Combined soil friction \& normal force per ubw. & Derived\\
        $F$ (\unit{\newton\per\meter}) & Tool force per ubw. & Derived\\
        $W$ (\unit{\newton\per\meter}) & Soil wedge weight per ubw. & Derived\\
        \midrule
        \multicolumn{3}{c}{FEE N Factors}\\
        \midrule
        $N_c$ (unitless) & Cohesion factor & Derived\\
        $N_{\gamma}$ (unitless) & Weight factor & Derived\\
        $N_Q$ (unitless) & Surcharge factor \& normal force per ubw. & Derived\\
        $N_{c_a}$ (unitless) & Adhesion factor & Derived\\
        \midrule

        \bottomrule
    \end{tabular}
    \label{tab:nomenclature}
\end{table}

Assuming equillibrium, i.e. no parts of the system are under acceleration, let us sum all of the forces in the $x$ and $z$ directions
\begin{flalign}
    \sum \mathbf{F}_x = 0 &= F \cos(\frac{\pi}{2}- \rho - \delta + \alpha) + c_a L_2 \cos(\rho - \alpha) - R \cos(\frac{\pi}{2} - \beta - \alpha - \phi) -c L_3 \cos(\alpha +\beta) &&\\
    \sum \mathbf{F}_z = 0 &= F \sin(\frac{\pi}{2}- \rho - \delta + \alpha) - c_a L_2 \sin(\rho - \alpha) + R \sin(\frac{\pi}{2} - \beta - \alpha - \phi) -c L_3 \sin(\alpha +\beta) - Q - W &&
\end{flalign}
Applying complementary angle rules \cite{joyce1996trig} and substituting for $L_2$ and $L_3$ yields
\begin{flalign}
    \sum \mathbf{F}_x = 0 &= F \sin(\rho + \delta - \alpha) + c_a d \frac{ \cos(\rho - \alpha)}{\sin(\rho)} - R \sin(\beta + \alpha + \phi) -c d \frac{\cos(\alpha +\beta)}{\sin(\beta)} &&\\
    \sum \mathbf{F}_z = 0 &= F \cos(\rho + \delta - \alpha) - c_a d \frac{\sin(\rho - \alpha)}{\sin(\rho)} + R \cos(\beta + \alpha + \phi) -c d \frac{\sin(\alpha +\beta)}{\sin(\beta)} - Q - W &&
\end{flalign}
Now linearly combine these equations to eliminate the $R$ terms
\begin{flalign}
    \sum \mathbf{F}_x \cot(\beta +\alpha + \phi) + \sum \mathbf{F}_z = 0 &= F \left( \cot(\beta +\alpha + \phi) \sin(\rho + \delta - \alpha) + \cos(\rho + \delta - \alpha)\right) && \nonumber \\
    &+\frac{c_a d}{\sin(\rho)} \left(\cos(\rho - \alpha) \cot(\beta +\alpha + \phi) - \sin(\rho - \alpha) \right) \nonumber\\
    &- \frac{c d}{\sin{\beta}} \left( \cos(\alpha +\beta) \cot(\beta +\alpha + \phi) + \sin(\alpha +\beta)\right) - Q - W \label{eq:FEE_inter1}
\end{flalign}
By multiplying Equation \ref{eq:FEE_inter1} by $\sin(\beta + \alpha + \phi)$ the $F$ term can be simplified using the Ptolemy's sum of angles identity in the following way
\begin{align*}
    &\sin(\beta + \alpha + \phi) \left[\cot(\beta +\alpha + \phi) \sin(\rho + \delta - \alpha) + \cos(\rho + \delta - \alpha) \right]\\
    &\Rightarrow \cos(\beta +\alpha + \phi) \sin(\rho + \delta - \alpha) + \sin(\beta + \alpha + \phi)\cos(\rho + \delta - \alpha)\\
    &\Rightarrow \sin(\beta + \phi + \rho + \delta)
\end{align*}
Assuming homogeneous soil, the weight of the soil wedge per ubw. can be expressed as the multiplication of the wedge cross-sectional area and the soil moist unit weight
\begin{align}
    W = \frac{1}{2} L_1 d \gamma = \frac{1}{2} d^2 ( \cot(\rho) + \cot(\beta)) \gamma \label{eq:FEE_wedge_weight}
\end{align}
Substituting \ref{eq:FEE_wedge_weight} and applying the simplification shown in Equation \ref{eq:FEE_inter1} the full expression then becomes
\begin{flalign}
    0 &= F \sin(\beta + \phi + \rho + \delta) + \frac{c_a d \sin(\beta + \alpha + \phi)}{\sin(\rho)} \left(\cos(\rho - \alpha) \cot(\beta +\alpha + \phi) - \sin(\rho - \alpha) \right) &&\nonumber\\
    &- \frac{c d \sin(\beta + \alpha + \phi)}{\sin{\beta}} \left( \cos(\alpha +\beta) \cot(\beta +\alpha + \phi) + \sin(\alpha +\beta)\right) && \nonumber\\
    &- \sin(\beta + \alpha + \phi)Q - \sin(\beta + \alpha + \phi)\frac{1}{2} d^2 ( \cot(\rho) + \cot(\beta)) &&
\end{flalign}
Solving for $F$
\begin{flalign}
    F &= - c_a d \frac{\sin(\beta + \alpha + \phi) \left(\cos(\rho - \alpha) \cot(\beta +\alpha + \phi) - \sin(\rho - \alpha) \right)}{\sin(\rho) \sin(\beta + \phi + \rho + \delta)} &&\nonumber\\
    &+ c d \frac{ \sin(\beta + \alpha + \phi)\left( \cos(\alpha +\beta) \cot(\beta +\alpha + \phi) + \sin(\alpha +\beta)\right)}{\sin(\beta)\sin(\beta + \phi + \rho + \delta})  && \nonumber\\
    &+ Q\frac{\sin(\beta + \alpha + \phi)}{\sin(\beta + \phi + \rho + \delta)}
    + d^2 \gamma \frac{\sin(\beta + \alpha + \phi) ( \cot(\rho) + \cot(\beta))}{2\sin(\beta + \phi + \rho + \delta)} &&
\end{flalign}
This expression then conforms to the standard FEE form where 4 terms are summed to represent the effects of of soil cohesion, weight, surcharge load, and soil-blade adhesion.
\begin{flalign}
    F = c d N_c + d^2 \gamma N_{\gamma} + Q N_Q +  c_a d N_{c_a} 
\end{flalign}
where the $N_c$ and $N_{c_a}$ factors can be simplified using Ptolemy’s identities
\begin{align}
    N_c &= \frac{\sin(\beta + \alpha + \phi) \left( \cos(\alpha +\beta) \cot(\beta +\alpha + \phi) + \sin(\alpha +\beta)\right)}{\sin(\beta) \sin(\beta + \phi + \rho + \delta)} \nonumber\\
    N_c &= \frac{ \cos(\alpha +\beta) \cos(\beta +\alpha + \phi) + \sin(\alpha +\beta)\sin(\beta + \alpha + \phi)}{\sin(\beta) \sin(\beta + \phi + \rho + \delta)} \nonumber\\
    N_c &= \frac{\cos(\phi)}{\sin(\beta) \sin(\beta + \phi + \rho + \delta)} \label{eq:FEE_N_c}
\end{align}
and 
\begin{align}
    N_{c_a} &= -\frac{\sin(\beta + \alpha + \phi) \left(\cos(\rho - \alpha) \cot(\beta +\alpha + \phi) - \sin(\rho - \alpha) \right)}{\sin(\rho) \sin(\beta + \phi + \rho + \delta)} \nonumber \\
    N_{c_a} &= -\frac{\cos(\rho - \alpha) \cos(\beta +\alpha + \phi) - \sin(\rho - \alpha) \sin(\beta + \alpha + \phi)}{\sin(\rho) \sin(\beta + \phi + \rho + \delta)} \nonumber \\
    N_{c_a} &= \frac{-\cos(\beta + \phi + \rho)}{\sin(\rho) \sin(\beta + \phi + \rho + \delta)}
\end{align}
Collecting these definitions in one location for convenience, the FEE for inclined terrain is
\begin{align} \label{eq:FEE}
    F &= f_{\text{FEE}}(d, \alpha, \rho, \beta, \phi, c, 
    \gamma, \delta, c_a, Q) \\
    &= \gamma d^2 N_\gamma + c d N_c + Q N_Q + c_a d N_a \nonumber
\end{align}
\begin{equation} \label{eq:FEE_coeff}
    \begin{matrix}
    N_\gamma = \frac{(\cot(\rho) + \cot(\beta)) \sin(\alpha + \phi + \beta)}{2 \sin(\delta + \rho + \phi + \beta)}, &
    N_Q = \frac{\sin(\alpha + \phi + \beta)}{\sin(\delta + \rho + \phi + \beta)} \\
    N_c = \frac{\cos(\phi)}{\sin(\beta)\sin(\delta + \rho + \phi + \beta)}, &
    N_{c_a} = \frac{-\cos(\rho + \phi + \beta)}{\sin(\rho)\sin(\delta + \rho + \phi + \beta)}
    \end{matrix}
\end{equation}

It is helpful here to note that if a flat terrain is assumed, $\alpha=0$, then the $N$ factors reduce to the model presented by \cite{Mckyes1989_ch6} (Equations 6.29-6.32)
\begin{equation*} 
    \begin{matrix}
    N_\gamma = \frac{(\cot(\rho) + \cot(\beta)) \sin( \phi + \beta)}{2 \sin(\delta + \rho + \phi + \beta)}, &
    N_Q = \frac{\sin(\phi + \beta)}{\sin(\delta + \rho + \phi + \beta)} \\
    N_c = \frac{\cos(\phi)}{\sin(\beta)\sin(\delta + \rho + \phi + \beta)}, &
    N_{c_a} = \frac{-\cos(\rho + \phi + \beta)}{\sin(\rho)\sin(\delta + \rho + \phi + \beta)}
    \end{matrix}
\end{equation*}

Note that Equation \ref{eq:FEE_coeff} still contains the unknown soil failure angle $\beta$.
\cite{Mckyes1965_ch3} explains that since soil frictional properties alone determine the shape of the failure surface during soil cutting, that the most likely failure surface does not depend upon the affect of soil cohesion or the surcharge.
Therefore, they reason that the best estimate for the failure angle $\beta^*$ should be the angle at which $N_\gamma$ is minimized, resulting in the path of least resistance for the soil to fail, i.e.
\begin{equation} \label{eq:min_N_gamma}
    \beta^* = \argmin_{\beta} N_\gamma
\end{equation}
The minimum will be located at the point where the partial derivative of $N_\gamma$ with respect to $\beta$ is zero and where the second derivative is positive. Using the quotient rule we can find the first derivative
\begin{align}
    \frac{\partial N_\gamma^{\text{num}}}{\partial \beta} &= u^\prime = -\csc^2(\beta) \sin(\alpha + \phi + \beta) + \left[ \cot(\rho) +\cot(\beta) \right] \cos(\beta + \alpha + \phi) \\
    \frac{\partial N_\gamma^{\text{den}}}{\partial \beta} &= v^\prime = 2 \cos(\delta + \rho +\phi + \beta) \\
    \frac{\partial N_\gamma}{\partial \beta} &= \frac{u^{\prime} v - u v^{\prime} }{v^2} = \frac{\frac{\partial N_\gamma^{\text{num}}}{\partial \beta} N_\gamma^{\text{den}} - N_\gamma^{\text{num}} \frac{\partial N_\gamma^{\text{den}}}{\partial \beta}}{N_\gamma^{\text{den}2}} \\
    &= \frac{\left[ \cot(\rho) + \cot(\beta)\right]sin(\delta + \rho - \alpha) -\csc^2(\beta) \sin(\alpha +\phi + \beta) \sin(\delta +\rho +\phi +\beta)}{2\sin^2(\delta+\rho+\phi+\beta)} = 0 \label{eq:FEE_dN_dbeta}
\end{align}
where one of Ptolmey's identites was used in the last line to simplify the expression.
It is not obvious how to solve Equation \ref{eq:FEE_dN_dbeta} to find a closed form expression for $\beta^*$.
Mckyes mentions that finding a closed form expression may not always be possible and that numerical methods may have to be relied upon \cite{Mckyes1989_ch6}.
However, for the flat terrain case $\alpha = 0$, a closed form solution for $\beta^*$ is provided (Equation 6.33 \cite{Mckyes1989_ch6}), although no derivation is given.
Miedema does derive an expression for $\beta^*$ for a smooth vertical blade moving through a flat cohesionless soil $(\delta=\SI{0}{\degree}, \rho=\SI{90}{\degree}, \alpha=\SI{0}{\degree}, c=0, c_a=0)$ \cite{miedema_2019} yielding
\begin{equation}
    \beta = \frac{\pi}{4}-\frac{1}{2}\phi
\end{equation}
but it isn't clear how to apply the same substitution ``trick'' to the more complex scenario.


The failure force can further be broken down into it's Cartesian components
\begin{equation}
    \mathbf{F} = \left[ F\cos(90\unit{\degree}-\rho-\delta+\alpha), F\sin(90\unit{\degree}-\rho-\delta+\alpha) \right]
\end{equation}

In order to handle the different scenarios of passive soil failure, when the soil wedge is moving up vs down with respect to the blade, the sign of the soil/tool friction and adhesion angles can be modified
\begin{equation}
\begin{matrix}
    \delta^\prime = \tanh(-C_1 \mathbf{v} \cdot \mathbf{i}_b)\delta, &
    c_a^\prime = \tanh(-C_1 \mathbf{v} \cdot \mathbf{i}_b) c_a \label{eq:FEE_vel_scaling}
\end{matrix}
\end{equation}
$\mathbf{v}$ is the vector velocity of the blade, $C_1$ is a scalar constant, and $\mathbf{i}_b$ is a unit vector pointing up along the blade \cite{Holz2013}.
The $\tanh(\cdot)$ here is used to smooth the transition between the two failure cases and avoid oscillations in the force when this model is used for simulation.
Note that this still assumes that the blade is moving forward through the soil and not in reverse which can occur in ``back-dragging" operations.

$0 < \delta + \rho + \phi + \beta < \pi$

Another complication/limitation of this model is that there is a singularity in the model when $\delta + \rho + \phi + \beta >= \pi$.
For illustrative purposes, assume no surcharge, adhesion, or cohesion.
This singularity then occurs when the soil failure force $R$ becomes co-linear with the blade force $F$ and the equilibrium conditions no longer hold.
CM Labs, the makers of the Vortex simulation engine, handle this by detecting this condition and assuming a high resistance reaction scaled linearly with the depth, presumably using a reaction force similar to
\begin{equation}
    F = C_2 (d +D_o) 
\end{equation} \label{eq:FEE_approx_singular}
where $C_2$ and $D_o$ are constants \cite{Holz2013}.

While not as accurate as some methods, the fundamental equation of earthmoving with coefficients derived for the trial wedge passive failure case (moving blade) provides a suitable approximation of soil mechanics for the purpose of earthmoving automation.
Both CM Labs \cite{Holz2013}and and recent research out of ETH Zurich \cite{Egli2022} use similar models to develop simulation of construction operations, indicating that these groups have reched similar conclusions.

\section{Vortex Soil-Tool Interaction Simulation}
The Vortex dynamics engine enables simulating of multi-body dynamics and soil-tool interaction.
In order to determine how this simulator can be utilized for earthmoving autonomy research, it is important to understand how the underlying physics of soil-tool interaction is simulated.
The first generation of this simulator is described in \cite{Holz2009, Holz2009a}, with subsequent improvements in \cite{Holz2013, Holz2014}.
Additional documentation is provided by CM Labs, the makers of Vortex Studio, in the Theory Guide \cite{cmlabs_2016} and basic simulator configuration guides \href{https://vortexstudio.atlassian.net/wiki/spaces/VSD225/pages/3250307323/Soil+Materials}{Soil Materials} and \href{https://vortexstudio.atlassian.net/wiki/spaces/VSD225/pages/3250294129/Earthwork+Systems+Tutorial+3+Soil+Parameters}{Earthworks Systems Tutorials 3: Soil Parameters}.

The soil is modeled using a hybrid heightfield-particle representation.
An FEE based model is used to determine the reaction force of the soil prior to shearing.
As the relative density $I_d$ of the soil increases, $\phi$ and $c$ are varied to increase the resistance of the soil to shearing due to stronger particle interlocking, and $\gamma$ is varied to adjust the weight of the soil wedge and surcharge \cite{Holz2009}.
When the force exerted by the tool on the soil reaches this threshold, the portion of the heightfield in contact with the blade is converted into particles whose behavior is governed by a discrete element model (DEM) simulation which more accurately characterizes the behavior of the disturbed soil.
As particles come to rest, they are then reintegrated back into the heightfield representation.
The initial relative density is defined as a fixed value but may vary throughout the simulation as the soil is compacted by the tool or as sheared particles are merged back into the heightfield \cite{Holz2009, Holz2013, Holz2014}.
The remainder of the soil properties remain constant for a given soil type and do not vary within a simulation episode. 

Further modifications to the Vortex soil simulation were carried out in \cite{Haeri2020}, including a term that reduces the contribution of the surcharge by a factor of 10 in Equation \ref{eq:FEE}.
Additionally, they find that selecting $\beta$ such that the top surface of the soil wedge matches extent of the surcharge pile better matches the experimental data as compared to minimization of $N_\gamma$ as in Equation \ref{eq:min_N_gamma}.
These changes enabled preditions of the soil-tool interaction forces for a lunar simulant material within 20\%-30\% of the measured force obtained from real-world experimentation.

\subsection{Accessing Vortex Soil Parameters}
\begin{figure}[t]
    \centering
    \begin{subfigure}[T]{0.3\textwidth}
        \centering
        \includegraphics[width=\textwidth]{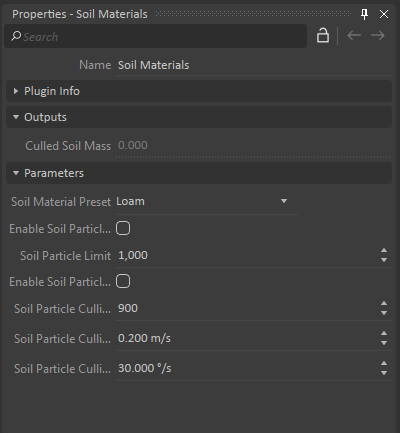}
        \caption{Default}
    \end{subfigure}
    \begin{subfigure}[T]{0.3\textwidth}
    \centering
        \includegraphics[trim={0 10cm 0 0}, clip, width=\textwidth]{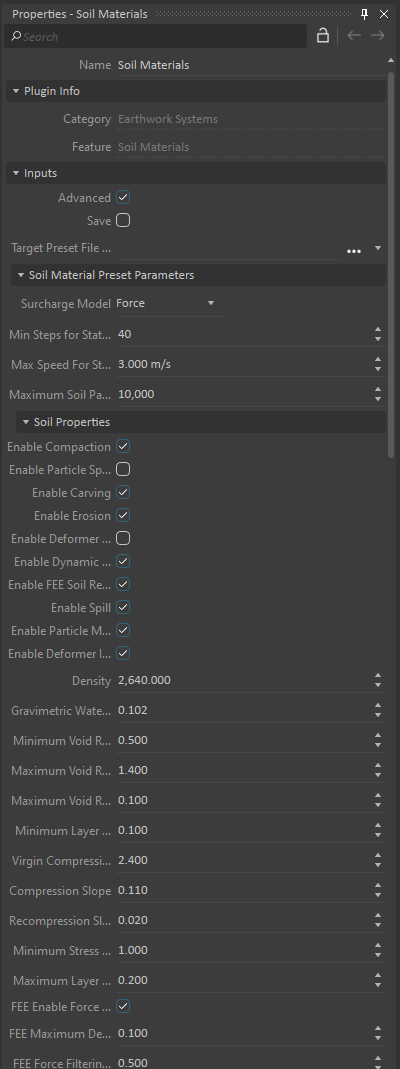}
        \caption{Advanced}
    \end{subfigure}
    \caption{Accessing advanced soil properties within Vortex Studio Soil Materials Extension}
    \label{fig:Vortex_soil_params}
\end{figure}
\subsubsection{Vortex Studio}
An \href{https://vortexstudio.atlassian.net/wiki/spaces/VSD238/pages/3342542865/Earthwork+Zone}{Earthworks Zone} or \href{https://vortexstudio.atlassian.net/wiki/spaces/VSD238/pages/3342542341/Deformable+Terrain}{Deformable Terrain} within a single simulation must be defined by a single soil type, i.e. multiple soil types within a single simulation is not supported.
There is no substantial difference in the physics between the Deformable Terrain and the Earthworks Zone.
The Deformable Terrain is meant to support large terrain areas by tiling Earthwork Zones internally and also enables use of more than one zone within a scene.
Vortex provides 4 default soil types: Clay, Loam, Sand, and Gravel. As new versions of Vortex Studio are released, sometimes these soils are updated as well. Older versions of these soils are also available for use.
These soil types are defined by .vxr (json) files typically contained within the following directory \path{C:\CM Labs\Vortex Studio 202X.Y\resources\EarthworkSystems\SoilMaterialPresets}.
Depending on the version of Vortex Studio, it is possible to access these soil parameters using a hidden menu:
\begin{enumerate}
    \item Click on Soil Materials extension (Likely under Earthworks Systems folder)
    \item Rotate view (Middle Click and Drag)
    \item Press the key combination ALT + Backspace. Then release
    \item Press a
\end{enumerate}
This will pull up an advanced view in the property viewer with many more parameters available as shown in Figure \ref{fig:Vortex_soil_params}.
The values here correspond to the values defined within the .vxr file corresponding to the soil type selected in the Soil Material Preset parameter.
If the steps above do not work to access the advanced menu, it may be necessary to load the scene using the Vortex Studio Python API, set the Advanced input flag to True, and save off the modified scene.

\subsection{Python API}
After loading the scene using the Vortex Python 3 API, a handle to the parameters of the Soil Materials extension can be obtained with
\begin{lstlisting}[language=Python]
soil_mat_parameter_container = scene.findExtensionByName("Soil Materials").getExtension().getParameterContainer()
\end{lstlisting}
Note that for any parameter changes to take effect, the Advanced flag must be set on the input container
\begin{lstlisting}[language=Python]
scene.findExtensionByName("Soil Materials").getInputContainer()['Advanced'].value = True
\end{lstlisting}

\subsubsection{Soil Type}
To select the soil type one must assign the correct integer value to the Soil Material Preset
\begin{lstlisting}[language=Python]
soil_mat_parameter_container['Soil Material Preset'].value = i
\end{lstlisting}
where $i$ corresponds to the index of the list specifying the available soil types. This is the default order that I believe the list is in, as it is the way these show up in Vortex studio.
\begin{lstlisting}[language=Python]
soil_types = ['Clay', 'Gravel', 'Loam', 'Sand',
              'Clay (2018a)', 'Gravel (2018a)', 'Loam (2018a)', 'Sand (2018a)',
              'Clay (2018b)', 'Gravel (2018b)', 'Loam (2018b)', 'Sand (2018b)',
              'DZ2019-Undisturbed_Generic', 'Loam_FEEv1'] 
\end{lstlisting}

\subsubsection{Advanced Soil Properties}
The advanced soil properties, e.g. FEE Tool Soil Adhesion ($c_a$), can be accessed using
\begin{lstlisting}[language=Python]
soil_prop = soil_mat_input_container["Soil Material Preset Parameters"]["Soil Properties"]
soil_tool_adhesion = soil_prop["FEE Tool Soil Adhesion"].value
\end{lstlisting}
Other parameters can be accessed similarly with the exception of Tables which are represented as a row-column list of lists.
For example the Internal Friction Coefficient Table can be converted to a numpy array 
\begin{lstlisting}[language=Python]
soil_friction_table_vx = soil_prop["Internal Friction Coefficient Table"]
nrows = len(soil_friction_table_vx)
soil_friction_table = np.zeros((nrows,2))
for row in range(nrows):
    soil_friction_table[row,0] = soil_friction_table_vx[row][0].value
    # The soil_friction_table_vx in Vortex stores tan(phi).
    soil_friction_table[row,1] = soil_friction_table_vx[row][1].value
\end{lstlisting}

\subsubsection{Initial Relative Density}
The initial relative density is not listed as a parameter of the Soil Materials extension and is instead specified in the Terrain Dynamics extension.
It can be accessed in the following manner
\begin{lstlisting}[language=Python]
ter_dyn_parameter_container = scene.findExtensionByName("Terrain Dynamics").getExtension().getParameterContainer()
ter_dyn_parameter_container['Relative Density'].value = init_density
\end{lstlisting}

\subsection{Soil Properties}\label{sec:vx_soil_prop}
The Vortex Studio soil simulation is complex, difficult to tune, and default soil properties are considered adequate for most end-user applications. 
The soil properties are normally only tuned by CM Labs or partner research institutions such as McGill and Concordia University, and therefore, there is little documentation on how each of these parameters actually affects the simulation beyond the source code.
However, for this research we require more detailed knowledge of the soil properties for validation of the soil property estimation system \cite{wagner2023}.
CM Labs has provided us with some limited internal documentation on the VxSoil API, which was helpful in understanding the role of some of these parameters, but is not publicly available.
This is not intended to provide comprehensive documentation of all Vortex Studio soil simulation.
The focus is primarily on understanding the soil properties related to the FEE portion of the model, however, other parameters may be discussed.
In particular, many of the parameters related to particles are not addressed.
The following sections are organized based on groupings of parameters that affect the same components of the soil model.

\subsubsection{Specifying Density of Soil} \label{sec:soil_density}
\begin{table}[h]
    \centering
    \begin{tabular}{l c  c} 
         \textbf{Vortex Parameter Name} & \textbf{Mathematical Expression}  & \textbf{Units}\\ [0.5ex] 
         \hline
         Relative Density & $I_d$ & \unit{\percent}\\ 
         \hline
         Void Ratio & $e$ & \text unitless\\
          \hline
         Minimum Void Ratio & $e_\text{min}$ & unitless\\
          \hline
         Maximum Void Ratio & $e_\text{max}$ & unitless \\
         \hline
         Gravimetric Water Content & $w$  & unitless\\
         \hline
         Density & $\rho_p$ & \unit{\kilo\gram\per\meter^3}
    \end{tabular}
    \caption{Table relating Vortex soil density parameter names to mathematical expressions. All Vortex parameters can be accessed through the \textit{Advanced Properties} in the \textit{Soil Materials} extension with the exception of \textit{Relative Density} which can be accessed through the \textit{Terrain Dynamics} extension.}
    \label{tab:soil_den}
\end{table}
There are a number of ways that can be used to express the density and/or compaction of a soil.
Within Vortex, the internal friction $\tan(\phi)$ and cohesion $c$ vary in a piece-wise linear fashion, see Section \ref{sec:vx_fee_soil_strength}, with respect to void ratio $e$ which is a measure of soil compaction computed as 
\begin{equation}
    e = \frac{V_V}{V_S}
\end{equation}
where $V_V$ and $V_S$ is the volume of voids and soils in a fixed total volume $V_T = V_V + V_S$ respectively \cite{Holz2009}.
Equivalently, the relative density of the soil expresses the compactness of a soil relative to the most loose $e_{\text{max}}$ and most compact $e_{\text{min}}$ state a soil can attain
\begin{equation}
    I_d = \frac{e_{\text{max}}-e}{e_{\text{max}} - e_{\text{min}}}*100
\end{equation}
Note, that the $e$ and $I_d$ are inversely related, e.g. a higher void ratio yields a lower relative density.

Vortex also enables varying soil initial relative density at different locations throughout the same simulation using the \href{https://vortexstudio.atlassian.net/wiki/spaces/VSD239/pages/3360364486/Earthwork+Utilities#Soil-Layer}{Soil Layer} extension.
This enables simulation of more complex soil conditions emulating stratification of soil commonly found in the real world.

To obtain a force from the FEE, the soil moist unit weight, $\gamma$ expressed in \unit{\newton\per\meter^3}, is required, see Equation \ref{eq:FEE}.
However, Vortex does not enable direct specification of $\gamma$.
Instead, the gravimetric water content $w$, soil particle density $\rho_\text{p}$, and initial relative density $I_d$ must be specified.
The gravimetric water content is the unitless ratio of water mass to soild mass for a soil, $w = m_w/m_s$.
The soil particle density is the density of the solid components of a soil and is expressed in \unit{\kilo\gram\per\meter^3}
Soil moist unit weight may be computed from these values as
\begin{equation}
    \gamma = \frac{\rho_\text{p}}{\rho_\text{w}}\gamma_w \frac{1+w}{1+e} \label{eq:moist_unit_weight}
\end{equation}
where $\rho_\text{w}=1000\unit{\kilo\gram\per\meter^3}$ is the density of water and $\gamma_w = 9810\unit{\newton\per\meter^3}$ is the specific weight of water.

\subsubsection{FEE Geometric Parameters} \label{sec:fee_geom}
\begin{table}[h]
    \centering
    \begin{tabular}{l c c } 
         \textbf{Vortex Parameter Name} & \textbf{Mathematical Expression} & \textbf{Units}\\ [0.5ex] 
         \hline
         FEE Slope Sample Cutoff Distance Factor & $d_\text{cutoff}$ & \unit{\meter}\\ 
         \hline
         FEE Slope Sample Point Weight Interpolation Coefficient & $\lambda$ & unitless\\
         \hline
         FEE Tool Wedge Contact Interpolation Coefficient & ? & ?
    \end{tabular}
    \caption{Table relating Vortex FEE geometric parameters names to mathematical expressions. These parameters can be accessed through the \textit{Soil Properties} in the \textit{Soil Materials} extension.}
    \label{tab:soil_geom}
\end{table}

\begin{figure}
    \centering
    \includegraphics[width=0.75\textwidth]{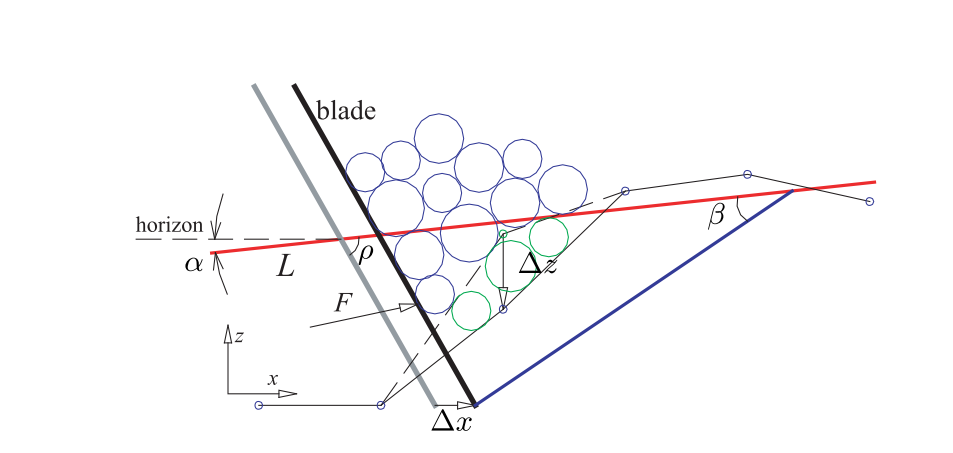}
    \caption{Figure reproduced from \cite{Holz2015} depicting the FEE slope determination and soil particle spawning processes. The red line indicates the approximated flat surface that is derived from the elevation map surface verticies. The thin dashed line indicates the surface prior to particle spawning and the thin solid line indicates the surface after spawning the new particles (shown in green).}
    \label{fig:fee_slope}
\end{figure}
The shape of the soil surface is used to compute the surface inclination and cut depth internally in Vortex.
The Vortex theory guide briefly talks about how this is accomplished \cite[p.~52]{cmlabs_2016}.
For a small width section of the blade, the soil height lying perpendicular to the blade is represented as a 1.5d elevation profile by sampling the 2.5d elevation map (which presumably involves some simple linear interpolation on the grid).
This profile must now be converted into a sloped flat surface to approximate the reaction force using the FEE, i.e. the green line in Figure \ref{fig:FEE_nomenclature}.
This is accomplished by computing a weighted linear least squares line fit on the profile where a point's weight is related exponentially from the distance to the tool \cite{Holz2013, Holz2015}.
No explicit expression is provided, however, based on this description we can surmise that the weighting takes a form similar to 
\begin{equation}
    W_{ii} = C_3\exp(-\lambda d), \quad d_i \in [0,d_\text{cutoff}]\unit{\meter}
\end{equation}
where $C_3$ and $\lambda$ are scaling constants, $d$ is the distance from the blade to the profile point, and $d_{cutoff}$ is the maximum distance between the blade and profile point for which a point will be included in computation of $W$.
My assumption is that $C_3$ is chosen to ensure that the weights sum to one
\begin{equation}
    C_3 = \frac{1}{N} \sum_{i=0}^{N-1}\exp{-\lambda d}
\end{equation}
This process is depicted in Figure \ref{fig:fee_slope}.

The weighted linear least squares for a line can be expressed in closed form as
\begin{align}
    \hat{y}_i &= f(x,\mathbf{a}) = a_0\phi_0(x_i) + a_1 \phi_1(x_i) = a_0 + a_1 x_i \\
    X_{ij} &= \phi_j(x_i) \\
    \hat{\mathbf{a}} &= (\mathbf{X}^\top \mathbf{W} \mathbf{X})^{-1} \mathbf{X}^\top \mathbf{W}\mathbf{z}
\end{align}
where $\mathbf{a}$ are the coefficients of the line, $\mathbf{X}$ is a matrix entries not defined above equal to zero, and $\mathbf{z}$ is the vector of profile heights.

The \textit{Advanced Soil Properties} contains two parameters that likely control this slope determination process: \textit{FEE Slope Sample Cutoff Distance Factor} and \textit{FEE Slope Sample Point Weight Interpolation Coefficient}.
The correspondence between these parameters and the above equations is shown in Table \ref{tab:soil_geom}
The line found using this procedure enables determination of a number of FEE geometric parameters: the blade angle wrt. the surface $\rho$, the surface angle $\alpha$, and cut depth perpendicular to the surface $d$.
This assumes knowledge of the orientation of the blade surface with respect to the soil surface.
For curved blades it is not obvious how this should be determined. 
This is discussed in \cite{Holz2013}, but instead of a height field, a signed distance field is used to represent the soil surface.
Through communications with CM Labs, it was determined that this is not actively being used as part of Vortex Studio as of the 2023.9 release.
It appears that by determining what parts of the blade are ``\textit{in contact}'' with the soil surface then a least squares fit can be used to approximate these points as a plane to determine the surface normal of the blade.
It is likely that the Vortex parameter \textit{FEE Tool Wedge Contact Interpolation Coefficient} controls the selection of points in the blade that are in contact with the soil in some manner.

\subsubsection{FEE Soil Strength Parameters} \label{sec:vx_fee_soil_strength}
\begin{table}[h]
    \centering
    \begin{tabular}{l c c } 
         \textbf{Vortex Parameter Name} & \textbf{Mathematical Expression} & \textbf{Units}\\ [0.5ex] 
         \hline
         Internal Friction Coefficient Table & $f_{\phi}(e)$ & unitless\\ 
         \hline
         Cohesion Table & $f_c(e)$ & \unit{\kilo\pascal}\\
         \hline
         FEE Tool Soil Friction Angle & $\delta$ & \unit{\radian} \\
         \hline
         FEE Tool Soil Adhesion & $c_a$ & \unit{\pascal}
    \end{tabular}
    \caption{Table relating Vortex FEE soil strength parameters names to mathematical expressions. These parameters can be accessed through the \textit{Soil Properties} in the \textit{Soil Materials} extension.}
\end{table}

The FEE is used to compute the force required to shear the soil for a given geometric configuration as determined by the procedure outlined in Section \ref{sec:fee_geom}.
There are a number of properties of the soil that affect the reaction force including: soil internal angle of friction $\phi$, soil cohesion $c$, soil-tool friction angle $\delta$, soil-tool adhesion $c_a$, and soil moist unit weight $\gamma$.
Additionally, the soil failure angle $\beta$ affects the reaction force, although this variable is not a soil property itself as it is typically found through a minimization procedure, see Equation \ref{eq:min_N_gamma}.
However, a study evaluating the accuracy of the Vortex studio soil simulation for a lunar simulant material found that choosing $\beta$ based on the extent of the surcharge pile was found to result in better agreement with experiments \cite{Haeri2020}.
It is not clear which procedure for finding $\beta$ is currently used within the software, but given that other modifications to the simulation introduced within this study are supported in Vortex Studio (surcharge contribution factor), it is likely that the surcharge extent method is used.
The soil moist unit weight $\gamma$ varies with void ratio $e$ according to Equation \ref{eq:moist_unit_weight} and is specified according to a number of other Vortex parameters discussed in Section \ref{sec:soil_density}.

Vortex requires specification of $phi$ and $c$ as a table relating void ratio $e$ to $\tan(\phi)$ and $c$ respectively.
\begin{align}
    \tan(\phi) &= f_{\phi}(e) \\
    c &= f_c(e)
\end{align}
where it is assumed that $f_{\phi}()$ and $f_c()$ are piece-wise linear functions specified by the \textit{Internal Friction Coefficient Table} and \textit{Cohesion Table} relating the soil properties to the void ratio $e$.

\subsubsection{FEE Hyperparameters \& Simulation Stability}
\begin{table}[h]
    \centering
    \begin{tabular}{l c c } 
         \textbf{Vortex Parameter Name} & \textbf{Mathematical Expression} & \textbf{Units}\\ [0.5ex] 
         \hline
         FEE Force Filtering Factor & $a$ & unitless\\ 
         \hline
         FEE Enable Force Keep Alive & ? & boolean\\
         \hline
         FEE Maximum Deformer Soil Separation for Force Keep Alive & ? & \unit{\meter}\\
         \hline
         FEE Sampling Resolution & - & number of timesteps\\
         \hline
         FEE Min Sampling Distance & - & \unit{\meter}\\
         \hline
         FEE Tangent Force Fade Out Coefficient & $C_1$ & unitless\\
         \hline
         FEE Force Keep Alive Override & - & boolean\\
         \hline
         FEE Force Keep Alive Enabled & - & boolean\\
         \hline
         Enable FEE Soil Reaction Force & - & boolean\\
         \hline
         Enable Carving & - & boolean\\
    \end{tabular}
    \caption{Table relating Vortex FEE soil parameters related to simulation stability to mathematical expressions. These parameters can be accessed through the \textit{Soil Properties} in the \textit{Soil Materials} extension with the exception of the \textit{FEE Force Keep Alive Override} and \textit{FEE Force Keep Alive Enabled} which can be accessed through the \textit{Tool Preset Parameters}.}
    \label{tab:soil_stabliity_hyper}
\end{table}
A number of hyperprameters for controlling the soil simulation are available.
Those related specifically to the FEE are discussed here while the others are discussed in the sections relevant to the components of the simulation that they control.
These parameters typically include the \textit{Enable} prefix in the parameter name.
The computation of the FEE reaction force can be disabled entirely using the \textit{Enable FEE Soil Reaction Force} if desired, although this does not have an obvious use case besides for debugging purposes.
Additionally, the anti-aliasing carving of the height grid, i.e. deformation from horizontal interaction of a deformer with the soil grid \cite{Holz2009}, can be disabled using \textit{Enable Carving}.
This functionality is different than the \textit{Enable Particle Spawning} which allows the height grid to be deformed as if the carved volume of material would be converted into particles, but does not actually spawn any particles.

In order to maintain a stable simulation, given the typical large simulation timestep of $1 / 60\unit{\second}$, Vortex developers have implemented a number of filters and methods for reducing oscillations induced by changes in contact between the soil and deformer, e.g. blade.
The sign of the frictional and cohesional forces applied to the blade, as modeled in the derivation of the FEE, depend on the direction of relative motion between the blade and the soil.
As this relative velocity changes, a rapid change in the reaction force is predicted by the FEE.
To smooth this transition, Holz et. al. uses Equation \ref{eq:FEE_vel_scaling}, which remaps $\delta$ and $c_a$ based on the magnitude of the relative velocity \cite{Holz2013}.
The coefficient $C_1$ is used to affect this mapping.

The FEE force produced by Equation \ref{eq:FEE} may be noisy due to noise in the determination of contact between the deformer and the soil and therefore the this force is filtered using what appears to be a low-pass filter
\begin{equation}
    \Tilde{F}_k = a \Tilde{F}_{k-1} + (1-a)F_k
\end{equation}
where $a$ is the filter smoothing factor related to the filter cutoff frequency $w_c$ via $a = \exp(-w_c T)$.
If we assume that the Vortex parameter \textit{FEE Force Filtering Factor} corresponds to $a$, then the default value of 0.5 in the available soils would imply $w_c = -\ln(0.5)60 = 41.6\unit{\hertz}$ which seems reasonable.

Although this has not been verified, the parameters \textit{FEE Sampling Resolution} and \textit{FEE Min Sampling Distance} appear to be a way of limiting the frequency at which the FEE calculation is performed.
\textit{FEE Sampling Resolution} is an integer value and may correspond to the minimum number of timesteps that must pass prior to updating the FEE estimated interaction force.
Similarly, \textit{FEE Min Sampling Distance} may require the deformer to have moved a minimum distance prior to computation of the interaction force.

The FEE force keep alive functionality is used to maintain the previously computed FEE force when contact between the deformer and the soil is not made or when the blade has become stationary, e.g. when the interaction force is not overcome by the deformer.
The \textit{FEE Enable Force Keep Alive} parameter enables this functionality and the \textit{FEE Maximum Deformer Soil Separation for Force Keep Alive} limits use of the keep alive force to when the blade remains near the soil surface.
It seems that this behavior can be overridden, possibly for debugging, by setting the parameter \textit{FEE Force Keep Alive Override} to true and setting the \textit{FEE Force Keep Alive Enabled} to the desired value, both of which are available in the \textit{Tool Preset Parameters} group.

\subsubsection{Particles}
The parallel particles technique that is used to perform this DEM simulation is documented in \cite{Holz2009, Holz2014}.
Particles are spawned as a result of the deformer engaging with the soil surface and FEE interaction force limits being exceeded.
These particles are then allowed to interact with each other, the soil surface, and the deformer.
The particles that accumulate in front of the cutting tool, i.e. the surcharge, is used to define the surcharge force $Q$ used in the FEE, Equation \ref{eq:FEE}.
When particles are determined to be stationary, they are merged back into the heightmap in a volume preserving manner.
If particles move outside of the simulated soil region on to rigid terrain, i.e. spilled, then they may be culled or removed from the simulation if desired.
The total number of particles may also be limited to limit the computational burden.
Two different soil particle generators are available: RB Generator and PB Generator.
It is not documented what the differences between them are, but if based on communications with CM Labs, it is believed that the RB is not used anymore due to speed concerns.
It is possible that the the PB generator is the parallel particles implementation \cite{Holz2014} and the RB represents an earlier version \cite{Holz2009}.

The particles have large number of parameters that define the physics of the DEM.
Additionally, a number of meta-parameters control the behavior of the particles as a group and include: \textit{Enable Particle Spawning}, \textit{Enable Particle Merging}, \textit{Enable Soil Particle Limit}, \textit{Enable Soil Particle Culling}, \textit{Soil Particle Culling Min Age}, \textit{Soil Particle Culling Max Linear Velocity}, \textit{Soil Particle Culling Max Anglular Velocity},\textit{Soil Particle Limit}, \textit{Enable Soil Particle Limit}, \textit{Particle Generator Type}, and \textit{Enable Spill}.

\subsubsection{Compaction and Erosion}
\begin{table}[h]
    \centering
    \begin{tabular}{l c c } 
         \textbf{Vortex Parameter Name} & \textbf{Mathematical Expression} & \textbf{Units}\\ [0.5ex] 
         \hline
         Enable Compaction & - & boolean\\
         \hline
         Virgin Compression Line Intercept & - & -\\
         \hline
         Compression Slope & - & -\\
         \hline
         Recompression Slope & - & -\\
         \hline
         Minimum Stress for Soil Compaction & - &-\\
         \hline
         Minimum Layer Height for Soil Compaction & - & -\\
         \hline
         Enable Erosion & - & boolean
    \end{tabular}
    \caption{Table relating Vortex soil compaction and erosion parameters to mathematical expressions. These parameters can be accessed through the \textit{Soil Properties} in the \textit{Soil Materials} extension.}
    \label{tab:soil_compact_eros}
\end{table}
The void ratio of the soil, a measure of soil compaction, is maintained at each grid location as a singly linked list representing a layered soil column \cite{Holz2009}.
Soil particles that are merged back in to the soil grid may will take on the compaction properties derived from the DEM. 
Additionally, compaction of the soil columns resulting from a deformer exerting a force in the vertical direction on the soil is accounted for.
This results the void ratios for the columns under the deformer being updated along with the height of the grid cell.
The reaction force produced by the terrain in the vertical direction is then applied to the blade to impede further downward movement of the deformer.
In the case of a tire, the interaction between the tire model and the soil grid can result in changing of soil compaction and height, e.g. \href{https://vortexstudio.atlassian.net/wiki/spaces/VSD239/pages/3360368655/Modular+Vehicles+Tutorial+5+Tire+Ruts}{Tire Ruts Tutorial}.
There is currently no similar support for tracks compacting soil, but it is possible to crudely emulate tracks attaching blade parts to the track bottom surface to enable some compaction effects as a tracked vehicle traverses the terrain.

Vortex parameters that control the vertical compaction of soil include: \textit{Enable Compaction, Virgin Compression Line Intercept, Compression Slope, Recompression Slope, Minimum Stress for Soil Compaction}, and \textit{Minimum Layer Height for Soil Compaction}.

In addition to deformation caused by the deformer (blade or bucket) by carving or compaction, the soil grid can be eroded due to the inability of the soil to support its own weight.
When unable to hold its weight, soil will slip until a stable slope is reached.
This process is documented in \cite{Holz2009} and can be enabled with the parameter \textit{Enable Erosion}.

\subsubsection{Soil Penetration: Deformer Contacts and the Material Table}
\begin{table}[h]
    \centering
    \begin{tabular}{l c c } 
         \textbf{Vortex Parameter Name} & \textbf{Mathematical Expression} & \textbf{Units}\\ [0.5ex] 
         \hline
         Enable Deformer Contacts & - & boolean\\ 
         \hline
         Enable Dynamic Contact Stiffness & - & boolean\\
         \hline
         Dynamic Stiffness Void Ratio Sample Depth & - & \unit{\meter}\\
         \hline
         Enable Deformer Interaction & - & boolean\\
         \hline
         Material Table & - & -
    \end{tabular}
    \caption{Table relating Vortex soil deformer contact parameters to mathematical expressions. These parameters can be accessed through the \textit{Soil Properties} in the \textit{Soil Materials} extension with the exception of the \textit{Material Table} which is a separate extension.}
    \label{tab:soil_deformer}
\end{table}

In addition to FEE derived cutting forces and forces imparted on the deformer by the soil particles, forces derived from contact between the deformer and the soil surface can be included by setting \textit{Enable Deformer Contacts} to true.
These forces are generated by determining the penetration between the objects, developing constraints to alleviate this interpenetration, and solving a mixed linear complementarity problem \cite[s.~2.5]{cmlabs_2016}.
Frictional forces can additionally be incorporated into these constraints.

The original purpose of the deformer contacts component of the simulation was to enable realistic placement of an earthmoving tool, e.g. excavator bucket, sinking into the terrain only if the soil strength could not support the weight of the tool.
This component of the model therefore serves the purpose of modeling the non-carving penetration of a tool into the terrain.
If \textit{Enable Dynamic Contact Stiffness} is set to False then the the properties in the materials table for the multi-body-dynamics portion of the simulation will be used to generate the contact force.
However, if \textit{Enable Dynamic Contact Stiffness} is set to True then the soil void ratio of the soil at a distance of \textit{Dynamic Stiffness Void Ratio Sample Depth} into the terrain (presumably the distance from the normal of the contact plane) will be used to compute the contact properties using the critical state model.
The contact stiffness then is tied to the elastic range of the soil at the current compaction level.
When soil compation compaction is additionally enabled, Vortex is able to simulate soil hardening effects on the soil and the increasing resistance of the soil to penetration.

As no documentation on how deformer contacts are combined with the FEE forces has been found, I was initially concerned that the accuracy of the soil carving process could be negatively affected as larger reaction forces would be generated than is predicted by the FEE alone.

To elucidate the relationship between contact forces between the blade and the soil surface as compared to the forces produced by the FEE component of the simulation, some experimentation was performed using the simplified blade environment described in \cite{wagner2023}.
Logging and graphing the \textit{Soil Cutting Force} outputs from the Blade extension and comparing those with the constraint force produced by the constraints that control the vertical position and horizontal velocity of the blade it was possible to determine the approximate function of these different soil simulation components along with how they interact when enabled simultaneously.

When setting \textit{Enable Deformer Contacts}, \textit{Enable Carving}, \textit{Enable FEE Soil Reaction Force}, and \textit{Enable Particle Spawning} all to false and then commanding the blade to a height below the soil surface with a small forward velocity,  no forces other than those required to accelerate the blade and overcome gravity are observed. 
However, by setting \textit{Enable Deformer Contacts} to true significant vertical and horizontal forces are produced.
This implies that the the multi-body dynamics engine is using the contact between the deformer (blade) object and the soil height grid object to generate non-negligible forces on the blade in this scenario.
When \textit{Enable Carving} is set to true, the terrain deforms as the blade sweeps through the soil.
Even though FEE forces are not being generated, the by enabling carving, the overlap between the blade and the soil surface is reduced significantly and the contact forces are reduced to values similar to the initial condition with some added low amplitude periodic noise.
This indicates that when carving is enabled the affect of deformer contact is minimal, meaning that the FEE generated forces still drive the the reaction forces applied to the blade.
Our concern about the possibility of deformer contacts causing significant non-realistic cutting forces are thus alleviated.

Setting \textit{Enable Deformer Contacts} may result is slightly more accurate estimation of cutting forces if no vertical penetration of the surface is being performed.
However, during dozing operations this is not realistic as the blade must be raised and lowered through each cut/fill pass.
It is therefore our recommendation to set \textit{Enable Deformer Contacts} and \textit{Enable Dynamic Contact Stiffness} to true for the most realistic bulldozing behavior.
Alternatively or additionally, penetration effects can be approximated with the inclusion of FEE move up and down forces, see Section \ref{sec:fee_heuristics}.

The parameter \textit{Enable Deformer Interaction} disables all interaction between the blade and the soil surface, but still allows for interaction between the blade and the particles.

\subsubsection{Heuristic Modifications to FEE}\label{sec:fee_heuristics}
\begin{table}[h]
    \centering
    \begin{tabular}{l c c } 
         \textbf{Vortex Parameter Name} & \textbf{Mathematical Expression} & \textbf{Units}\\ [0.5ex] 
         \hline
         FEE Max Force Per Submerged Tool Area & $C_6$ &  \unit{\newton\per\meter^3}\\ 
         \hline
         FEE Move Down Reaction Force Per Submerged Tool Area & $C_4$ & \unit{\newton\per\meter^3}\\ 
         \hline
         FEE Move Up Reaction Force Per Submerged Tool Area& $C_5$ & \unit{\newton\per\meter^3}\\
         \hline
         Enable FEE Soil Reaction Force & - & boolean\\
         \hline
         Force Scale Table & $f_F(e)$ & unitless\\
         \hline
         FEE Force Scale Override & - & boolean\\
         \hline
         FEE Force Scale & - & boolean \\
         \hline
         FEE Surcharge Contribution Factor & $s_Q$ & unitless
    \end{tabular}
    \caption{Table relating heuristic Vortex FEE soil parameters to mathematical expressions. These parameters can be accessed through the \textit{Soil Properties} in the \textit{Soil Materials} extension with the exception of the \textit{FEE Force Scale Override} and \textit{FEE Force Scale} which can be accessed through the \textit{Tool Preset Parameters}.}
\end{table}
The FEE is a semi-empirical model that produces an interaction force necessary to shear a uniform soil with a flat blade.
The model simplifies the the soil-tool interaction significantly.
For example, no consideration of the force required for a tool penetrating a soil is considered.
The Vortex soil simulation introduces a number of heuristic modifications to the FEE to enable more realistic behavior of the system.
I have not be able to verify the exact purpose of the following parameters, but this is a hypothesis about how they affect the interaction force.
The \textit{FEE Move Down Reaction Force Per Submerged Tool Area} and \textit{FEE Move Up Reaction Force Per Submerged Tool Area} add additional vertical reaction forces on the deformer as it penetrates or raises through the soil
\begin{equation}
    F^\prime_z = F_z + F^\text{down}_z +F^{\text{up}}_z= F_z + C_4 A_s \mathbb{I}_{\left\{\mathbf{\overrightarrow{v}} \cdot \mathbf{\overrightarrow{i}_z} < 0\right\}} + C_5 A_s \mathbb{I}_{\left\{\mathbf{\overrightarrow{v}} \cdot \mathbf{\overrightarrow{i}_z} > 0\right\}}
\end{equation}
where $A_s$ is the submerged area, $C_4$ is the \textit{FEE Move Down Reaction Force Per Submerged Tool Area}, and $\mathbb{I}$ is an indicator function.
It is not entirely clear how $A_s$ is computed and if this is the projection of the submerged tool along the $xy$ plane or something else.
It is hard to determine without access to the code, but it appears that there is some sort of filter on the velocity condition that causes the additional forces to be applied after a short period of movement.
The parameter \textit{Enable FEE Soil Reaction Force} determines if these additional forces are applied or not.
Overall, these forces may help to simulate frictional and penetration forces better than what is accounted for simply by the deformer contact forces.

The force produced by the FEE module may be limited by setting \textit{FEE Max Force Per Submerged Tool Area}, denoted as $C_6$ or it may be scaled with void ratio $e$
\begin{align}
    F^\prime &= \text{sign}(F)\max(|F|,C_6)\\
    F^\prime &= f_F(e) F
\end{align}
where $f_F()$ is a piece-wise linear function specified by the \textit{FEE Force Scale Table} parameter.
There does not seem to be any documented soil mechanics motivated reason for this, but CM Labs appears use this functionality to tune soils for different earthmoving machines.
It appears possible to override this scaling by setting the \textit{FEE Force Scale Override} to true and setting the \textit{FEE Force Scale} to the desired value, both of which are available in the \textit{Tool Preset Parameters} group.

The contribution of the surcharge to the FEE may be reduced by modifying Equation \ref{eq:FEE} in the following manner
\begin{equation}
    F = \gamma d^2 N_\gamma + c d N_c + {\color{orange}s_Q} Q N_Q + c_a d N_a
\end{equation}
where $s_q \in [0,1]$ is the \textit{FEE Surcharge Contribution Factor}.
Tuning of $s_q$ was found to lead to better agreement between the Vortex soil model and experimental results \cite{Haeri2020}.

\subsection{Tuning for FEE-Only Behavior}
For the initial development and validation of of the FEE based physics infused neural network soil property estimation system a dataset of soil-tool interaction with simplified physics derived primarily from the FEE is required, referred to as $\mathcal{D}_\text{FEE}$ \cite{wagner2023}.
Given our understanding of the Vortex soil parameters documented in Section \ref{sec:vx_soil_prop}, the simulator was configured in the following way to achieve this:
\begin{table}[h]
    \centering
    \begin{tabular}{l c c } 
         \textbf{Vortex Parameter Name} & \textbf{Value}\\ [0.5ex]
         \hline
         FEE Enable Particle Spawning & False\\ 
         \hline
         FEE Enable Deformer Contacts & False\\ 
         \hline
         FEE Move Down Reaction Force Per Submerged Tool Area & 0.0\unit{\newton\per\meter^3}\\ 
         \hline
         FEE Move Up Reaction Force Per Submerged Tool Area & 0.0\unit{\newton\per\meter^3}\\
         \hline
         FEE Slope Sample Cutoff Distance Factor & 2.0\unit{\meter}\\ 
         \hline
         FEE Slope Sample Point Weight Interpolation Coefficient & 0.0\\
         \hline
         FEE Surcharge Contribution Factor & 1.0\\
         \hline
         FEE Tool Soil Friction Angle & 0.268\unit{\radian} (15\unit{\degree})
    \end{tabular}
    \caption{Values of Vortex soil property parameters modified from default in order to achieve a simplified FEE-only simulation.}
    \label{tab:fee_dataset}
\end{table}

Particle spawning is disabled so that forces exerted by the particles on the blade are not present.
All penetration effects are disabled including deformer contact and FEE move up and down forces.
The slope determination calculation is modified to enforce an assumption of flat terrain.
Setting $d_\text{cutoff}$ to a large number and $\lambda=0.0$ ensures that a large number of surface points are sampled with equal weight yielding a near 0 slope line.
The points nearest the blade will have a lower elevation as a result of carving being enabled, but this should only have a small effect.
Carving is left enabled to allow for the vehicle chassis to ride on the deformed surface, adding an element of realism to the dataset while retaining the ability to have ground truth knowledge of the soil properties.
The $s_q$ is set to 1 although this is not stritly necessary as particle spawning has been disabled.
Because the blade angle is fixed at $10\unit{\degree}$ and the default value for the soil-tool friction angle is also $10\unit{\degree}$ the vertical forces generated by the FEE are 0.
Therefore, $\delta$ is modified to $15\unit{\degree}$ just to have the simulation produce non-zero vertical forces.

While collecting of this dataset some odd behaviors were noted.
In general there is still some uncertainty in the ground-truth depth of cut.
This is likely partially related to the fact that internal to Vortex the slope of the terrain is being estimated with a slightly positive value due to deformation of the soil near the blade due to carving.
In validation of the model we assume that $\alpha=0$.
Additionally, some error may account for the fact that the blade itself is defined with a rectangular profile.
Given our uncertainty in how Vortex determines contact between the blade and soil, it is not clear whether the front edge of the blade or the lowest point is used to determine the depth of cut $d$.

In addition to the FEE dataset $\mathcal{D}_\text{FEE}$, a default dataset $\mathcal{D}_\text{default}$ is collected.
In order to accurately monitor surcharge $Q$ particle merging is disabled by setting \textit{Enable Particle Merging} to false. 
During data collection for $\mathcal{D}_\text{default}$ the blade was unable to penetrate the ground for the sand and gravel materials when the initial relative density was set to a high value, $I_d\approx60-100\unit{\percent}$.
The blade instead rode slid somewhat roughly along the flat surface generating few if any particles.
Initially we believed that this was a result of Vortex considering the bottom of the blade as the cutting surface when determining the blade angle $\rho$.
Setting the blade geometry thickness to a small value did help reduce the inability of the blade to penetrate the dense soil somewhat, but it was still evident at the highest $I_d$.
After some further experimentation, we now believe that this behavior is the result of sand and gravel having very small \textit{Compression Slope} values.
This small value means that the soil does not deform substantially even when large vertical loads are applied.
In this case this renders the blade unable to carve out any soil because of its inability to penetrate the surface.
We have not tested digging in a compacted cohesionless material, but this may actually be fairly accurate behavior.
If deformer contact is disabled by setting \textit{Enable Deformer Contacts} to False then the blade has no resistance to penetration.
This behavior is certainly not realistic, however tuning of the \textit{FEE Move Down Reaction Force Per Submerged Tool Area} may be able to capture some of the penetrative force effects.
If deformer contacts are enabled though this movce down force heuristic should likely be disabled to avoid accounting for penetrative effects with multiple simulation components.
This is something to consider when tuning a simulation for soil tool interaction realism.

\printbibliography

@incollection{Mckyes1989_ch2,
author = {Mckyes, E},
booktitle = {Agricultural Engineering Soil Mechanics},
doi = {10.1016/B978-0-444-88080-2.50007-7},
pages = {12--58},
title = {{Chapter 2. Soil Shear Strength}},
year = {1989}
}

@incollection{Mckyes1989_ch6,
author = {Mckyes, E},
booktitle = {Developments in Agricultural Engineering},
doi = {10.1016/B978-0-444-88080-2.50011-9},
issn = {01674137},
pages = {137--171},
title = {{Chapter 6. Lateral Earth Pressures}},
year = {1989}
}

@incollection{Mckyes1965_ch3,
author = {Mckyes, E},
booktitle = {Soil Cutting and Tillage},
doi = {10.1016/B978-0-444-42548-5.50006-4},
pages = {38--86},
title = {{Chapter 3. Soil Cutting Forces}},
year = {1965}
}

@article{Egli2022,
author = {Egli, Pascal ; and Gaschen, Dominique ; and Kerscher, Simon ; and Jud, Dominic ; and Hutter, Marco},
doi = {10.3929/ethz-b-000557541},
journal = {IEEE Robotics and Automation Letters},
title = {{Soil-Adaptive Excavation Using Reinforcement Learning}},
url = {https://doi.org/10.3929/ethz-b-000557541},
year = {2022}
}

@online{joyce1996trig,
  author = {David E. Joyce},
  title = {Summary of trigonometric identities},
  url = {https://www2.clarku.edu/faculty/djoyce/trig/identities.html},
  year = {1996},
  month = {9},
  day = {22},
  accessed = {2023-09-22},
}

@book{miedema_2019,
  title = {The Delft Sand, Clay \& Rock Cutting Model},
  author = {Miedema, Sape A.},
  publisher = {Delft University of Technology},
  year = {2019},
  chapter = {2.10: Passive Soil Failure},
  url = {https://eng.libretexts.org/Bookshelves/Civil_Engineering/Book%3A_The_Delft_Sand_Clay_and_Rock_Cutting_Model_(Miedema)/02%3A_Basic_Soil_Mechanics/2.10%3A_Passive_Soil_Failure}
}

@phdthesis{Holz2009a,
author = {Holz, Daniel},
title = {{Deformable Terrain for Real-Time Simulation}},
year = {2009}
}

@inproceedings{Holz2009,
author = {Holz, Daniel and Beer, Thomas and Kuhlen, Torsten},
booktitle = {VRIPHYS 2009 - 6th Workshop on Virtual Reality Interactions and Physical Simulations},
pages = {21--30},
title = {{Soil deformation models for real-time simulation: A hybrid approach}},
year = {2009}
}

@inproceedings{Holz2013,
author = {Holz, Daniel and Azimi, Ali and Teichmann, Marek},
booktitle = {Proceedings of the 30th IAARC International Symposium on Automation and Robotics in Construction (ISARC)},
title = {{Real-Time Simulation of Mining and Earthmoving Operations: A Level Set-Based Model for Tool-Induced Terrain Deformations}},
year = {2013}
}

@online{cmlabs_2016,
  author = "CM Labs Simulations",
  title = "Theory guide: Vortex software's multibody dynamics engine",
  year = "2016",
  url = "https://vortexstudio.atlassian.net/wiki/spaces/ VSD21A/pages/2823143940/Vortex+Theory+Guide+Document",
  accessed = "2023-03-01"
}

@inproceedings{wagner2023,
  title={In Situ Soil Property Estimation for Autonomous Earthmoving Using Physics-Infused Neural Networks},
  author={Wagner, W Jacob and Soylemezoglu, Ahmet and Nottage, Dustin and Driggs-Campbell, Katherine},
  booktitle={Proceedings of the 16th European-African Regional Conference of the ISTVS},
  year={2023}
}

@inproceedings{wagner2025,
  title={In-situ soil-property estimation and {Bayesian} mapping with a simulated compact track loader},
  author={Wagner, W Jacob and Soylemezoglu, Ahmet and Driggs-Campbell, Katherine},
  booktitle={Proceedings of the 55th Conference of the ISTVS},
  year={2025}
}

@article{Haeri2020,
author = {Haeri, A. and Tremblay, D. and Skonieczny, K. and Holz, D. and Teichmann, M.},
journal = {Proceedings of the 37th International Symposium on Automation and Robotics in Construction, ISARC 2020: From Demonstration to Practical Use - To New Stage of Construction Robot},
number = {December},
pages = {608--615},
title = {{Efficient Numerical Methods for Accurate Modeling of Soil Cutting Operations}},
year = {2020}
}

@inproceedings{Holz2014,
    author = {Holz, D.},
    title = {Parallel Particles (P2): A Parallel Position Based Approach for Fast and Stable Simulation of Granular Materials},
    booktitle = {VRIPHYS 2014 - 11th Workshop on Virtual Reality Interactions and Physical Simulations},
    pages = {135--144},
    year = 2014
}

@inproceedings{Holz2015,
author = {Holz, Daniel and Azimi, Ali and Teichmann, Marek},
booktitle = {2015 International Conference on Virtual Reality and Visualization},
pages = {166--172},
title = {{Advances in physically-based modeling of deformable soil for real-time operator training simulators}},
year = {2015}
}

\end{document}